\documentclass{llncs}
\usepackage{graphicx}
\usepackage{multicol}
\usepackage{amsmath}
\usepackage{makeidx}
\usepackage{subcaption}

\begin{document}
%
%
\title{Object-Oriented Knowledge Representation and Data Storage Using Inhomogeneous Classes}
\titlerunning{Object-Oriented Knowledge Representation and Data Storage Using Inhomogeneous Classes}  
%
\author{Dmytro Terletskyi}
\authorrunning{} 
%
\tocauthor{}
\institute{Taras Shevchenko National University of Kyiv, Kyiv, 03680, Ukraine
\email{dmytro.terletskyi@gmail.com},\\
}

\maketitle              

\begin{abstract}
This paper contains analysis of concept of a class within different object-oriented knowledge representation models. The main attention is paid to structure of the class and its efficiency in the context of data storage, using object-relational mapping. The main achievement of the paper is extension of concept of homogeneous class of objects by introducing concepts of single-core and multi-core inhomogeneous classes of objects, which allow simultaneous defining of a few different types within one class of objects, avoiding duplication of properties and methods in representation of types, decreasing sizes of program codes and providing more efficient information storage in the databases. In addition, the paper contains results of experiment, which show that data storage in relational database, using proposed extensions of the class, in some cases is more efficient in contrast to usage of homogeneous classes of objects.

\keywords{class, homogeneous class, single-core inhomogeneous class, core of level $m$, multi-core inhomogeneous class}
\end{abstract}

\section{Introduction}

During recent years amount of information, which is used by information systems, has extremely increased, therefore invention of efficient approaches to knowledge representation, storing and extraction of data become more crucial. Nowadays the majority of modern information systems is developed using object-oriented programming (OOP) languages. The main advantage of OOP-languages is simultaneous combining of the programming paradigm implementation and object-oriented knowledge representation model within the language. Moreover, almost all modern languages support such technology as object-relational mapping (ORM), which allows data storing within relational database in terms of classes and objects.

However, one of the important tasks of object-oriented information systems development, is design of classes and their hierarchy, which describes particular domain. Complexity of the whole system mostly depends on the structure of the classes and the structure of their hierarchy. In addition, efficiency of data storage within relational database also depends on the structure of particular classes.

Nowadays concept of a \emph{class} is widely used in almost all approaches to knowledge representation. However, its definition and interpretations within various knowledge representation formalisms have some differences. Let us consider the notion of the class within such object-oriented knowledge representation models (KRMs) as \emph{frames} and \emph{object-oriented programming} (OOP).

\section{Concept of a Class}

In the theory of frames \emph{class} is usually called \emph{class-frame} or \emph{generic frame} \cite{Minsky}, \cite{Negnevitsky}, \cite{Brachman-Levesque}. \emph{Class-frame} is defined as a data structure, which describes a group of objects with common attributes \cite{Negnevitsky}. It defines attributes (properties), relationships with other frames and procedural attachments (methods) for class-instances or other class-frames. In other words, a class-frame defines a stereotypical object, which can be considered as a template for creation of more specific stereotypical or particular objects. Therefore, all frames, which are instances of a class-frame, have the same structure and behavior.

Within the OOP concept of the \emph{class} is associated with a number of different, but not always competing, interpretations \cite{Craig}, \cite{Booch}, \cite{Meyer}, \cite{Dathan-Ramnath}, \cite{Weisfeld}. It can variously be interpreted as: a set of objects; a program structure or module; a factory-like entity which creates objects; a data type. However, despite all differences, a class defines some set of properties (specification) and (or) set of methods (signature), which are common for all objects of the class, in all of these interpretations. In other words, specification defines the structure of objects, while signature defines their behavior.

Analyzing and comparing notions of a class within the theory of frames and OOP, it is possible to conclude that in both cases a class defines objects with same structure and behavior, i.e. objects of the same type. Therefore, such classes can be called \emph{homogeneous} ones. This feature of classes introduces some limitation, in particular it makes impossible simultaneous definition of a few types of objects within one class. Consequently, description of each new type of objects requires definition of new class. When a few types of objects are similar but not equivalent ones, it can cause duplication of properties and methods in classes, which describe these types. To prevent such situations, frames, as OOP, support the inheritance mechanism, that helps to build class hierarchies in more flexible and efficient way. However, as it was shown in \cite{Touretzky}, \cite{Al-Asady}, \cite{Terletskyi-1}, using of inheritance can cause problems of exceptions, redundancy and ambiguity, which usually arise while constructing of hierarchies and reasoning over them.

\section{Homogeneous Classes of Objects}

Besides frames and OOP, there is such object-oriented knowledge representation model as \emph{object-oriented dynamic networks} (OODN), which was proposed in \cite{Terletskyi-2}, \cite{Terletskyi-3}, \cite{Terletskyi-4}. All these KRMs have some similarity, however OODN use more extended notion of the class, than frames and OOP, which allows avoiding duplication of properties and methods in representation of types, decreasing sizes of program codes and providing more efficient information storage in the databases. Let us consider concept of class within OODN in more details.

Similarly to frames and OOP, OODN exploit concept of \emph{homogeneous class} (HC) of objects.
\begin{definition}
\label{HC_def}
Homogeneous class of objects $T$ is a tuple $T=(P(T),F(T))$, where $P(T)=(p_1(T),\dots,p_n(T))$ is a specification, which defines some quantity of objects, and $F(T)=(f_1(T),\dots,f_m(T))$ is a signature, which can be applied to them.
\end{definition}
As it was mentioned early, such concept of the class has widespread practical usage in frames systems and OOP, however there are some objects that simultaneously belong to many different classes, which cannot be described using concept of homogeneous classes. One of the approaches for solving this problem is introduction of concept of \emph{inhomogeneous class} of objects, which extends notion of \emph{homogeneous class} \cite{Terletskyi-2}, \cite{Terletskyi-5}. However, concept of inhomogeneous classes, which is proposed in \cite{Terletskyi-2}, can be extended by the dividing of inhomogeneous classes of objects on \emph{single-core inhomogeneous classes} (SCIC) and \emph{multi-core inhomogeneous classes} (MCIC) of objects. Let us define these kinds of inhomogeneous classes of objects.

\section{Single-Core Inhomogeneous Classes of Objects}

Concept of single-core inhomogeneous class of objects means the same as concept of homogeneous class of objects, which was defined in \cite{Terletskyi-2}, i.e.
\begin{definition}
\label{SCIC_def}
Single-core inhomogeneous class of objects $T$ is a tuple
\[T=(Core(T),pr_1(A_1),\dots,pr_n(A_n)),\]
where $Core(T)=(P(T),F(T))$ is a core of the class $T$, which contains properties and methods that are common for objects $A_1,\dots,A_n$, and $pr_i(A_i)=(P(A_i),F(A_i))$, where $i=\overline{1,n}$, is an $i$-th projection of the class $T$, which contains properties and methods that are typical only for the object $A_i$, $i=\overline{1,n}$.
\end{definition}
As we can see, concept of SCIC allows describing two or more different types within one class, using OOP-like style, while describing each new type within OOP requires definition of new class or using mechanism of inheritance if types have common properties and (or) methods.

Analyzing Def.~\ref{HC_def} and Def.~\ref{SCIC_def}, we can conclude that any HC defines only one type of objects, while any SCIC defines at least two different types of objects. Therefore, in the first case notions of \emph{class} and \emph{type} mean the same, while in second case they have different meaning. Taking into account that SCIC simultaneously defines a few types of objects, let us introduce the concept of a \emph{type} of objects.
\begin{definition}
\label{type_def}
Type $t_i$, $i=\overline{1,n}$ of inhomogeneous class of objects $T_{t_1,\dots,t_n}$ is a homogeneous class of objects $t_i=(Core(T_{t_1,\dots,t_n}),pr_i(A_i))$, where $Core(T_{t_1,\dots,t_n})$ is a core of the class $T_{t_1,\dots,t_n}$, and $pr_i(A_i)$ is its $i$-th projection.
\end{definition}

Let us consider an example of SCIC of objects.
\begin{example}
\label{sc_inh_class_example}
Clearly that such geometric figures as square, rectangle and rhombus belong to the class of convex quadrangles. Let us define SCIC of objects $T_{SRRb}$, which defines these types of convex quadrangles in the following way
\begin{align*}
T_{SRRb}=(p_1(T_{SRRb})&=(4,\text{sides}),\\
p_2(T_{SRRb})&=(4,\text{angles}),\\
p_3(T_{SRRb})&=vf_3(T_{SRRb})=(1),\\
f_1(T_{SRRb})&=(v_1(p_2(t_i)+v_2(p_2(t_i))+v_3(p_2(t_i))+v_4(p_2(t_i)),\text{cm}),\\
&\ \ \ \ i\in\{S,R,Rb\}\\
p_1(t_S)&=((2,\text{cm}),(2,\text{cm}),(2,\text{cm}),(2,\text{cm})),\\
p_2(t_S)&=((90^\circ),(90^\circ),(90^\circ),(90^\circ)),\\
p_3(t_S)&=vf_3(t_S)=(1),\\
p_4(t_S)&=vf_4(t_S)=(1),\\
f_1(t_S)&=\left((v_1(p_1(t_S)))^2,\text{cm}^2\right),\\
p_1(t_R)&=((2,\text{cm}),(3,\text{cm}),(2,\text{cm}),(3,\text{cm})),\\
p_2(t_R)&=((90^\circ),(90^\circ),(90^\circ),(90^\circ)),\\
p_3(t_R)&=vf_3(t_R)=(1),\\
p_4(t_R)&=vf_4(t_R)=(1),\\
f_1(t_R)&=\left(v_1(p_1(t_R))\cdot v_2(p_1(t_R)),\text{cm}^2\right),\\
p_1(t_{Rb})&=((3,\text{cm}),(3,\text{cm}),(3,\text{cm}),(3,\text{cm})),\\
p_2(t_{Rb})&=((80^\circ),(100^\circ),(80^\circ),(100^\circ)),\\
p_3(t_{Rb})&=vf_3(t_{Rb})=(1),\\
p_4(t_{Rb})&=vf_4(t_{Rb})=(1),\\
f_1(t_{Rb})&=\left.\left((v_1(p_1(t_{Rb})))^2\cdot\sin(v_1(p_4(t_{Rb}))),\text{cm}^2\right)\right),
\end{align*}
where $p_1(T_{SRRb})$ is a quantity of sides, $p_2(T_{SRRb})$ is a quantity of internal angles, $vf_3(T_{SRRb})$ is a verification function, which defines a property ``sum of all internal angles is equal to $360^\circ$'', i.e. $vf_3(T_{SRRb}):p_3(T_{SRRb})\rightarrow\{0,1\}$, where
\[p_3(T_{SRRb})=(v_1(p_4(t_i))+v_2(p_4(t_i))+v_3(p_4(t_i))+v_4(p_4(t_i))=360),\]
where $i\in\{S,R,Rb\}$, $f_1(T_{SRRb})$ is a method of perimeter calculation, $p_1(t_S)$, $p_1(t_R)$, $p_1(t_{Rb})$ are sizes of sides, $p_2(t_S)$, $p_2(t_R)$, $p_2(t_{Rb})$ are degree measures of internal angles, $vf_3(t_S)$ is a verification function, which defines a property ``all sides of figure have the same length'', i.e. $vf_3(t_S):p_3(t_S)\rightarrow\{0,1\}$, where
\[p_3(t_S)=(v_1(p_1(t_S))=v_2(p_1(t_S))=v_3(p_1(t_S))=v_4(p_1(t_S))),\]
$vf_4(t_S)$ is a verification function, which defines a property ``all internal angles are equal to $90^\circ$'', i.e $vf_4(t_S):p_4(t_S)\rightarrow\{0,1\}$, where
\[p_4(t_S)=(v_1(p_2(t_S))=v_2(p_2(t_S))=v_3(p_2(t_S))=v_4(p_2(t_S))=90),\]
$f_1(t_S)$ is a method of square calculation, $vf_3(t_R)$ is a verification function, which defines a property ``opposite sides of the figure have the same length'', i.e. $vf_3(t_R):p_3(t_R)\rightarrow\{0,1\}$, where
\[p_3(t_R)=((v_1(p_1(t_R))=v_3(p_1(t_R)))\wedge(v_2(p_1(t_R))=v_4(p_1(t_R)))),\]
$vf_4(t_R)$ is a verification function, which defines a property ``all internal angles are equal to $90^\circ$'', i.e $vf_4(t_R):p_4(t_R)\rightarrow\{0,1\}$, where
\[p_4(t_R)=(v_1(p_2(t_R))=v_2(p_2(t_R))=v_3(p_2(t_R))=v_4(p_2(t_R))=90),\]
$f_1(t_R)$ is a method of square calculation, $vf_3(t_{Rb})$ is a verification function, which defines a property ``all sides of figure have the same length'', i.e. $vf_3(t_{Rb}):p_3(t_{Rb})\rightarrow\{0,1\}$, where
\[p_3(t_{Rb})=(v_1(p_1(t_{Rb}))=v_2(p_1(t_{Rb}))=v_3(p_1(t_{Rb}))=v_4(p_1(t_{Rb}))),\]
$vf_4(t_{Rb})$ is a verification function, which defines a property of equality of opposite internal angles of the figure, i.e. $vf_4(t_{Rb}):p_4(t_{Rb})\rightarrow\{0,1\}$, where
\[p_4(t_{Rb})=((v_1(p_2(t_{Rb}))=v_3(p_2(t_{Rb})))\wedge(v_2(p_2(t_{Rb}))=v_4(p_2(t_{Rb})))),\]
³ $f_1(t_{Rb})$ is a method of square calculation.

As we can see, SCIC of objects $T_{SRRb}$ simultaneously describes three types of convex quadrangles $t_S$, $t_R$ and $t_{Rb}$. Therefore, concept of SCIC of objects allows describing of classes, which define two and more types of objects. Such approach gives us an opportunity of efficient knowledge representation due construction of core of inhomogeneous class of objects.

Indeed, from the described example, we can see that for representation of types, which define squares, rectangles and rhombuses, it is necessary to describe $7$ properties and $2$ methods for each type, i.e. $21$ properties and $6$ methods. Usage of the SCIC provides representation of these types via representation of only $3$ properties and $1$ method for the class core, and $4$ properties and $1$ method for each of projections of the class, i.e. $15$ properties and $4$ methods. In such a way, proposed approach allows avoiding duplication of properties and methods in representation of types, decreasing sizes of program codes and providing more efficient information storage in the databases.
\end{example}

\section{Multi-Core Inhomogeneous Classes of Objects}

According to Def.~\ref{SCIC_def}, core of the class contains only properties and methods, which are common for all types of the class and projections of the class contain properties and methods, which are typical only for the particular types. However, sometimes a few projections can contain equivalent properties and (or) methods, which are typical not for all types of the class, therefore they are not parts of the class core. In these cases duplication of such properties and (or) methods will occur. In order to prevent it and to make the class structure more optimal, let us define the concept of \emph{core of level $m$}.
\begin{definition}
\label{core_level_m_def}
Core of level $m$ of inhomogeneous class $T_{t_1,\dots,t_n}$ is a tuple
\[Core^m\left(T_{t_1,\dots,t_n}\right)=\left(P\left(T_{t_{i_1},\dots,t_{i_m}}\right),F\left(T_{t_{i_1},\dots,t_{i_m}}\right)\right),\]
where $t_{i_1},\dots,t_{i_m}$ are arbitrary $m$ types from the set of types $\{t_1,\dots,t_n\}$, where $1\leq m\leq n$, $1\leq i_1\leq\dots\leq i_m\leq n$, and $P\left(T_{t_{i_1},\dots,t_{i_m}}\right)$, $F\left(T_{t_{i_1},\dots,t_{i_m}}\right)$ are specification and signature of the core of inhomogeneous class $T_{t_{i_1},\dots,t_{i_m}}$, which contain properties and methods, which are common for all objects of types $t_{i_1},\dots,t_{i_m}$.
\end{definition}

Since, not all types of the class can have common properties and (or) methods, the inhomogeneous class of objects, which defines $n$ types, can contain $k$ cores of level $m$, where $0\leq k\leq C_n^m$. That is why, let us generalize the Def.~\ref{SCIC_def}, taking into account Def.~\ref{type_def} and Def.~\ref{core_level_m_def}.
\begin{definition}
\label{MCIC_def}
Multi-core inhomogeneous class of objects $T_{t_1,\dots,t_n}$ is a tuple
\begin{gather*}
T_{t_1,\dots,t_n}=\left(Core^n_1(T_{t_1,\dots,t_n}),Core^{n-1}_1(T_{t_1,\dots,t_n}),\dots,Core^{n-1}_{k_{n-1}}(T_{t_1,\dots,t_n}),\dots,\right.\\
\left.Core^1_1(T_{t_1,\dots,t_n}),\dots,Core^1_{k_1}(T_{t_1,\dots,t_n}),pr_1(t_1),\dots,pr_n(t_n)\right),
\end{gather*}
where $Core^n_1(T_{t_1,\dots,t_n})$ is a core of level $n$ of the class $T_{t_1,\dots,t_n}$, $Core^{n-1}_{i_{n-1}}(T_{t_1,\dots,t_n})$ is an $i_{n-1}$-th core of level $n-1$ of the class $T_{t_1,\dots,t_n}$, where $i_{n-1}=\overline{1,k_{n-1}}$ and $k_{n-1}\leq C_n^{n-1}$, $Core^1_{i_1}(T_{t_1,\dots,t_n})$ is an $i_1$-th core of level $1$ of the class $T_{t_1,\dots,t_n}$, where $i_1=\overline{1,k_1}$ and $k_1\leq C_n^1$, $pr_i(t_i)$ is an $i$-th projection of the class $T_{t_1,\dots,t_n}$, which contains properties and methods, which are typical only for the type $t_i$, where $i=\overline{1,n}$.
\end{definition}

Let us consider an example of MCIC of objects.
\begin{example}
\label{mc_inh_class_example}
Let us consider all types of convex quadrangles from the previous example and define MCIC of objects $T_{SRRb}$, which defines all these types in the following way
\begin{align*}
T_{SRRb}=(p_1(T_{SRRb})&=(4,\text{sides}),\\
p_2(T_{SRRb})&=(4,\text{angles}),\\
p_3(T_{SRRb})&=vf_3(T_{SRRb})=(1),\\
f_1(T_{SRRb})&=(v_1(p_2(t_i))+v_2(p_2(t_i))+v_3(p_2(t_i))+v_4(p_2(t_i)),\text{cm}),\\
&\ \ \ \ i\in\{S,R,Rb\}\\
p_1(T_{SR})&=((90^\circ),(90^\circ),(90^\circ),(90^\circ)),\\
p_2(T_{SR})&=vf_2(t_i)=(1),\ i\in\{S,R\}\\
p_1(T_{SRb})&=vf_1(t_i)=(1),\ i\in\{S,Rb\}\\
f_1(T_S)&=\left((v_1(p_1(t_S)))^2,cm^2\right),\\
p_1(T_R)&=vf_1(t_R)=(1),\\
f_1(T_R)&=\left(v_1(p_1(t_R))\cdot v_2(p_1(t_R)),\text{cm}^2\right),\\
p_1(T_{Rb})&=vf_1(t_{Rb})=(1),\\
f_1(T_{Rb})&=\left((v_1(p_1(t_{Rb})))^2\cdot\sin(v_1(p_4(t_{Rb}))),\text{cm}^2\right),\\
\end{align*}
\begin{align*}
p_1(t_S)&=((2,\text{cm}),(2,\text{cm}),(2,\text{cm}),(2,\text{cm})),\\
p_1(t_R)&=((2,\text{cm}),(3,\text{cm}),(2,\text{cm}),(3,\text{cm})),\\
p_1(t_{Rb})&=((3,\text{cm}),(3,\text{cm}),(3,\text{cm}),(3,\text{cm})),\\
p_2(t_{Rb})&=((80^\circ),(100^\circ),(80^\circ),(100^\circ))),
\end{align*}
where $p_1(T_{SRRb})$ is a quantity of sides, $p_2(T_{SRRb})$ is a quantity of internal angles, $vf_3(T_{SRRb})$ is a verification function, which defines a property ``sum of all internal angles is equal to $360^\circ$'', i.e. $vf_3(T_{SRRb}):p_3(T_{SRRb})\rightarrow\{0,1\}$, where
\[p_3(T_{SRRb})=(v_1(p_4(t_i))+v_2(p_4(t_i))+v_3(p_4(t_i))+v_4(p_4(t_i))=360),\]
where $i\in\{S,R,Rb\}$, $f_1(T_{SRRb})$ is a method of perimeter calculation, $p_1(T_{SR})$ is degree measures of internal angles, $vf_2(T_{SR})$ is a verification function, which defines a property ``all internal angles are equal to $90^\circ$'', i.e. $vf_2(T_{SR}):p_2(T_{SR})\rightarrow\{0,1\}$, where
\[p_2(T_{SR})=(v_1(p_1(T_{SR}))=v_2(p_1(T_{SR}))=v_3(p_1(T_{SR}))=v_4(p_1(T_{SR}))=90),\]
where $i\in\{S,R\}$, $vf_1(T_{SRb})$ is a verification function, which defines a property ``all sides of figure have the same length'', i.e. $vf_1(T_{SRb}):p_1(T_{SRb})\rightarrow\{0,1\}$, where
\[p_1(T_{SRb})=(v_1(p_1(t_i))=v_2(p_1(t_i))=v_3(p_1(t_i))=v_4(p_1(t_i))),\ i\in\{S,Rb\},\]
$f_1(T_S)$ is a method of square calculation, $vf_1(t_R)$ is a verification function, which defines a property ``opposite sides of the figure have the same length'', i.e. $vf_1(t_R):p_1(T_R)\rightarrow\{0,1\}$, where
\[p_1(t_R)=((v_1(p_1(t_R))=v_3(p_1(t_R)))\wedge(v_2(p_1(t_R))=v_4(p_1(t_R)))),\]
$f_1(T_R)$ is a method of square calculation, $vf_1(t_{Rb})$ is a verification function, which defines a property of equality of opposite internal angles of the figure, i.e.   $vf_1(t_{Rb}):p_1(T_{Rb})\rightarrow\{0,1\}$, where
\[p_1(t_{Rb})=((v_1(p_2(t_{Rb}))=v_3(p_2(t_{Rb})))\wedge(v_2(p_2(t_{Rb}))=v_4(p_2(t_{Rb})))),\]
$f_1(T_{Rb})$ is a method of square calculation, $p_1(t_S)$, $p_1(t_R)$, $p_1(t_{Rb})$ are sizes of sides,
$p_2(t_{Rb})$ are degree measures of internal angles.

Analyzing the structure of the class $T_{SRRb}$, we can see that it is really multi-core, because it has $1$ core of level $3$, $2$ cores of level $2$ and $3$ cores of level $1$. The structures of all cores of the class $T_{SRRb}$ are shown in the Table~\ref{tab-1}.
\begin{table}
\caption{Structures of cores of the class $T_{SRRb}$}
\label{tab-1}
\begin{center}
\renewcommand{\arraystretch}{1.4}
\setlength\tabcolsep{3pt}
\begin{tabular}{lll}
\hline\noalign{\smallskip}
\multicolumn{1}{c}{\bf Core} & \multicolumn{1}{c}{\bf Properties / Methods} & \multicolumn{1}{c}{\bf Common for types} \\
\noalign{\smallskip}
\hline
\noalign{\smallskip}
$Core^3_1(T_{SRRb})$ & $p_1(T_{SRRb})$, $p_2(T_{SRRb})$, $p_3(T_{SRRb})$, $f_1(T_{SRRb})$ & $t_S$, $t_R$, $t_{Rb}$\\
$Core^2_1(T_{SRRb})$ & $p_1(T_{SR})$, $p_2(T_{SR})$ & $t_S$, $t_R$\\
$Core^2_2(T_{SRRb})$ & $p_1(T_{SRb})$ & $t_S$, $t_{Rb}$\\
$Core^1_1(T_{SRRb})$ & $f_1(T_S)$ & $t_S$\\
$Core^1_2(T_{SRRb})$ & $p_1(T_R)$, $f_1(T_R)$ & $t_R$\\
$Core^1_3(T_{SRRb})$ & $p_1(T_{Rb})$, $f_1(T_{Rb})$ & $t_{Rb}$\\
\hline
\end{tabular}
\end{center}
\end{table}

As we can see, MCIC of objects $T_{SRRb}$ simultaneously describes three types of convex quadrangles $t_S$, $t_R$ and $t_{Rb}$ too. Therefore, concept of MCIC of objects also allows describing of classes, which define two and more types of objects. Such approach gives us an opportunity of efficient knowledge representation due construction of cores of level $m$ of inhomogeneous class of objects.

From the previous example, it is known that for representation of types, which define squares, rectangles and rhombuses, it is necessary to describe $7$ properties and $2$ methods for each type, i.e. $21$ properties and $6$ methods. Usage of the MCIC provides representation of these types via representation of only $3$ properties and 1 method for the $Core^3_1(T_{SRRb})$, $2$ properties for the $Core^2_1(T_{SRRb})$, $1$ property for the $Core^2_2(T_{SRRb})$, $1$ method for the $Core^1_1(T_{SRRb})$, $1$ property and $1$ method for the $Core^1_2(T_{SRRb})$, $1$ property and $1$ method for the $Core^1_3(T_{SRRb})$, $1$ property for the $pr_1(t_S)$, $1$ property for the $pr_1(t_R)$ and $2$ properties for the $pr_1(t_{Rb})$, i.e. $12$ properties and $4$ methods. In such a way, usage of MCIC similarly to SCIC allows avoiding duplication of properties and methods in representation of types, decreasing sizes of program codes and provides more efficient information storage in the databases.
\end{example}

\section{Object-Oriented Data Storage}

Nowadays almost all object-oriented information systems use the databases for data storage, therefore efficient representation of classes and objects within the databases (in particular within relational databases) and their further extraction are topical issues. Modern stack of object-oriented technologies contains such tool as object-relational mapping (ORM), which allows mapping of classes and their objects into tables and their records within relation databases \cite{Ambler}, \cite{Philippi}, \cite{Torres-Galante}. Despite all advantages of ORM, it has some limitations, in particular it still does not provide an ability to map methods of classes into the database that makes impossible precise and complete exchange of classes among different object-oriented software. Moreover, as it was emphasized in \cite{Goncalves}, inheritance is not a feature that relational databases naturally have, and therefore the mapping is
not as obvious.

It is known that during the ORM structures of classes define the structures of future tables in the database. Therefore efficiency of data storage in the database strictly depends on structures of classes. That is why we performed an experiment for checking, how the usage of SCICs and MCICs influences representation of objects and their types in relational databases in contrast to usage of HCs.

As a part of the experiment, we have created three relational databases using HCs, SCICs and MCICs. We used types $t_S$, $t_R$ and $t_{Rb}$ form the Example~\ref{sc_inh_class_example} and Example~\ref{mc_inh_class_example} as a test data.

The database based on HCs has $3$ tables, where each of them has $31$ columns for every type of objects, i.e. $t_S$, $t_R$ and $t_{Rb}$. The database based on SCICs has $4$ tables: $1$ table with the $9$ columns for the $Core(T_{SRRb})$ and $3$ tables with the $23$ columns for $pr_1(S)$, $pr_2(R)$, $pr_3(Rb)$. The database based on MCICs has $9$ tables: $1$ table with $9$ columns for the $Core^3_1(T_{SRRb})$, $1$ table with $11$ columns for the $Core^2_1(T_{SRRb})$, $1$ table with $3$ columns for the $Core^2_2(T_{SRRb})$, $1$ table with $3$ columns for the $Core^1_1(T_{SRRb})$, $1$ table with $5$ columns for the $Core^1_2(T_{SRRb})$, $1$ table with $5$ columns for the $Core^1_3(T_{SRRb})$ and three tables which have $9$, $11$ and $19$ columns respectively, for $pr_1(S)$, $pr_2(R)$ and $pr_3(Rb)$.

The experiment was performed in the environment of operating system Linux Debian Stretch. MariaDB~10.1.23 was chosen as a DB server. The aim of the experiment was to compare sizes of DBs, which are deployed on the server, and sizes of their exported *.sql files. Corresponding measurements were performed for $21$ cases. First measurement was done when DBs contained $0$ objects, after this we added $4000$ objects of each types (i.e. $12000$ objects during one insertion session) to every DB and repeated the measurements. Then the procedure was repeated $19$ times. At the end of the experiment each DB contained $80000$ objects of each types (i.e. $240000$ objects in each DB). For simplification of automated database generation, all objects of the same type were initialized in the same way. Results of all measurements are shown in the Tab.~\ref{tab-2}.
\begin{table}[t!]
\caption{Sizes of DBs based on HCs, SCICs, MCICs and their exported *.sql files}
\label{tab-2}
\begin{center}
\renewcommand{\arraystretch}{1.4}
\setlength\tabcolsep{3pt}
\begin{tabular}{c|ccc|ccc}
\hline\noalign{\smallskip}
{\bf Quantity} & \multicolumn{3}{c|}{{\bf Sizes of DB (Mb)}} & \multicolumn{3}{c}{{\bf Sizes of DB *.sql file (Mb)}}\\ \cline{2-7}
{\bf of objects} & {\bf HC  } & {\bf SCIC } & {\bf MCIC } & {\bf HC} & {\bf SCIC  } & {\bf MCIC }\\
\noalign{\smallskip}
\hline
\noalign{\smallskip}
$0$ & $0.046875$ & $0.0625$ & $0.125000$ & $0.006396$ & $0.006080$ & $0.007398$\\
$12000$	& $4.546875$ & $4.5625$ & $1.031250$ & $4.730189$ & $2.954480$ & $0.872142$\\
$24000$	& $10.546875$ & $7.5625$ & $2.593750$ & $9.456926$ & $5.905566$ & $1.738131$\\
$36000$	& $13.546875$ & $8.5625$ & $4.640625$ & $13.747144$ & $8.861661$ & $2.610227$\\
$48000$	& $17.546875$ & $11.5625$ & $4.640625$ & $18.928142$ & $11.825956$ & $3.488397$\\
$60000$	& $21.546875$ & $14.5625$ & $5.640625$ & $23.666618$ & $14.789045$ & $4.366487$\\
$72000$	& $26.546875$ & $17.5625$ & $5.640625$ & $28.405355$ & $17.751929$ & $5.244479$\\
$84000$	& $30.546875$ & $20.5625$ & $8.640625$ & $33.143831$ & $20.715019$ & $6.122471$\\
$96000$	& $33.546875$ & $21.5625$ & $8.640625$ & $37.882568$ & $23.678108$ & $7.000561$\\
$108000$ & $40.546875$ & $23.5625$ & $8.640625$ & $42.621044$ & $26.641197$ & $7.878553$\\
$120000$ & $45.546875$ & $26.5625$ & $8.640625$ & $47.359781$ & $29.604286$ & $8.756643$\\
$132000$ & $45.546875$ & $29.5625$ & $9.640625$ & $52.098518$ & $32.567581$ & $9.634635$\\
$144000$ & $47.546875$ & $29.5625$ & $9.640625$ & $56.837220$ & $35.530940$ & $10.512725$\\
$156000$ & $47.546875$ & $29.5625$ & $9.640625$ & $61.575731$ & $38.493554$ & $11.390717$\\
$168000$ & $47.546875$ & $29.5625$ & $9.640625$ & $66.314207$ & $41.456849$ & $12.268807$\\
$180000$ & $47.546875$ & $29.5625$ & $13.640625$ & $71.053206$ & $44.419733$ & $13.146799$\\
$192000$ & $68.640625$ & $43.5625$ & $15.640625$ & $75.791682$ & $47.382822$ & $14.024791$\\
$204000$ & $75.640625$ & $47.5625$ & $15.640625$ & $80.530419$ & $50.346117$ & $14.902881$\\
$216000$ & $75.640625$ & $47.5625$ & $16.640625$ & $85.268895$ & $53.309206$ & $15.780873$\\
$228000$ & $84.656250$ & $52.5625$ & $16.640625$ & $90.007632$ & $56.272295$ & $16.659141$\\
$240000$ & $84.656250$ & $52.5625$ & $16.640625$ & $94.746369$ & $59.235385$ & $17.537246$\\
\hline
\end{tabular}
\end{center}
\end{table}

\begin{figure}[h!]
 \centering
 \includegraphics[width=1\textwidth]{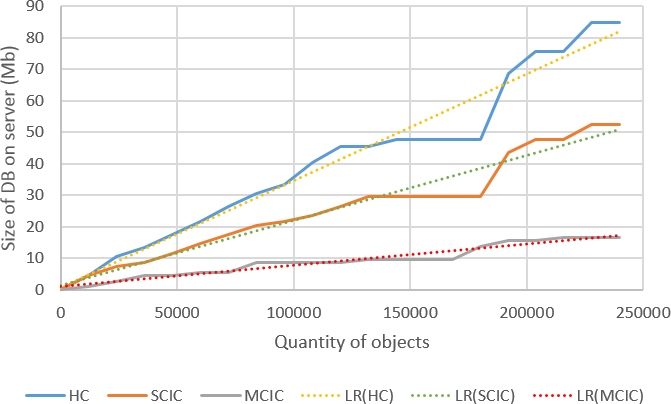}
 \caption{Comparison of sizes of DBs based on HCs, SCICs and MCICs of objects}
 \label{fig-1}
\end{figure}

Using obtained results, we built following dependencies between sizes of DBs and quantities of objects, which they contain
\begin{align}
 S(DB_{HC})&=0.0003*Q+0.793,\\
 S(DB_{SCIC})&=0.0002*Q+1.253,\\
 S(DB_{MCIC})&=0.00007*Q+1.0651,
\end{align}
where $S(DB_i)$ is a size of DB of $i$-th type, where $i\in\{HC,SCIC,MCIC\}$, and $Q$ is a quantity of objects within the DB. Corresponding graphs and their linear approximations are shown on Fig.~\ref{fig-1}.

\begin{figure}[h!]
 \centering
 \includegraphics[width=01\textwidth]{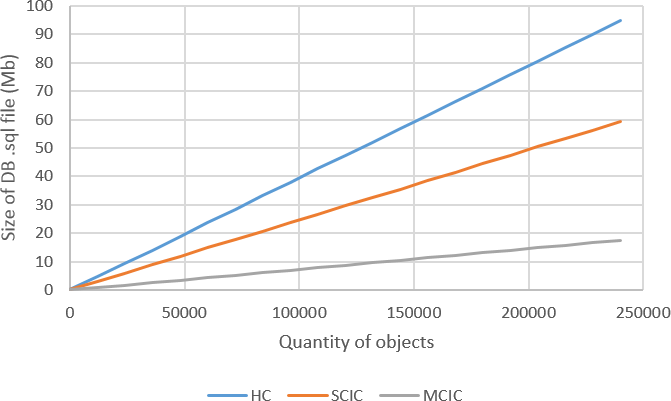}
 \caption{Comparison of sizes of exported *.sql files of DBs based on HCs, SCICs and MCICs of objects}
 \label{fig-2}
\end{figure}

In addition, we built following dependencies between sizes of exported *.sql files of DBs and quantities of objects within DBs:
\begin{align}
 S(FDB_{HC})&=0.0004*Q-0.0781,\\
 S(FDB_{SCIC})&=0.0002*Q-0.0.171,\\
 S(FDB_{MCIC})&=0.00007*Q-0.0145,
\end{align}
where $S(FDB_i)$ is a size of exported *.sql DB file of $i$-th type, where $i\in\{HC,SCIC,MCIC\}$, and $Q$ is a quantity of objects within the DB. Corresponding graphs and their linear approximations are shown on Fig.~\ref{fig-2}.

Obtained dependencies allow us to predict approximate size of DB  and its exported *.sql file, based on quantity of objects within the environment where experiment was performed. However, these results strictly depend on the environment of experiment, in particular type of operating system, DB server, database engine, parameters of DB tables, etc. Nevertheless, they show the efficiency of using MCIC comparing with using SCIC and HC.

For convenient usage of proposed approach, we can formulate the following theorem.
\begin{theorem}
\label{MCIC_efficiency}
Efficiency of storage $m_1,\dots,m_n$ objects of types $t_1,\dots,t_n$ in relation database, using concept of MCIC in contrast to usage of HC can be calculated in the following way
\[E=100-\frac{M_{MCIC}}{M_{HC}}\cdot100,\]
where
\[M_{MCIC}=\left(\sum_{i_n=1}^1C^n_1+\sum_{i_{n-1}=1}^{k_{n-1}}C^{n-1}_{i_{n-1}}+\dots+\sum_{i_1=1}^{k_1}C^1_{i_1}+\sum_{i=1}^nPr_i\cdot m_i\right),\]
and $C^n_{i_n},\dots,C^1_{i_1}$~-- are memory sizes, which allow storing of $i_n,\dots,i_1$ cores of level $n,\dots,1$ of class $T_{t_1,\dots,t_n}$ respectively, $Pr_i$~-- is memory size, which allows storing of $i$-th projection of the class, and $M_{HC}=T_1\cdot m_1+\dots+T_n\cdot m_n$, where $T_i$~-- is memory size, which allows storing of $i$-th type of the class.
\end{theorem}

Analyzing the theorem, let us formulate a few important remarks.
\begin{remark}
\label{MCIC_efficiency_r1}
Precise calculation of efficiency of data storage in relation database, using concept of MCIC, before creation of the database itself and its information filling, is impossible, because the storing of equivalent properties for objects of the same type can require different memory sizes.
\end{remark}
\begin{remark}
\label{MCIC_efficiency_r2}
Th.~\ref{MCIC_efficiency} allows calculating approximate efficiency coefficient of data storage in relation database, using concept of MCIC, before (without) the database creation. It can be done, if maximum possible memory sizes, which allow storing properties and methods of each type, are used for calculation.
\end{remark}
\begin{remark}
\label{MCIC_efficiency_r3}
Th.~\ref{MCIC_efficiency} can be easily reformulated for calculation of efficiency coefficient of using MCIC comparing with using SCIC by applying memory sizes, which allow storing SCICs instead $T_1,\dots,T_n$.
\end{remark}

Summarizing results of performed experiment, it is possible to conclude that in the case of using MCICs:
\begin{enumerate}
 \item storage of $n$ objects, where $n\in\{0,12000, 24000,\dots, 240000\}$, in the DB is more efficient in average by $64.93\%$ in contrast to using of SCICs and $77.89\%$ in contrast to using of HCs;
 \item size of exported *.sql file of such DB decreased in average by $70.41\%$ in contrast to using of SCICs and $81.5\%$ in contrast to using of HCs;
 \item SQL queries to tables of such DB are performed faster in the contrast to cases of using SCICs and HCs.
\end{enumerate}

Analyzing Def.~\ref{SCIC_def} and Def.~\ref{MCIC_def}, it is possible to conclude, that usage of MCIC or SCIC is efficient, when types of objects has common properties and (or) methods. In such cases structure of the MCICs will be similar to classes hierarchies built using rational single inheritance, which allow avoiding of the redundancy problem. However, MCICs can be used even in situations when types of objects do not have common properties and (or) methods, in these cases classes will have only projections, which are equivalent to HCs.

\section{Conclusions and Outlook}

In this paper we analized concept of a class within such object-oriented KRMs as frames, OOP and OODN. The main attention was paid to structure of the class and its efficiency in the context of data storage, using object-relational mapping.

The main achievement of the paper is introduction of concepts of single-core and multi-core inhomogeneous classes of objects, which extend notion of homogeneous class of objects. The main idea of SCIC is defining single core of level $n$, while the main idea of MCIC is defining set of cores of level $m$. Proposed concepts allow simultaneous defining of a few different types within one class of objects, avoiding duplication of properties and methods in representation of types (especially if they have common properties and (or) methods), decreasing sizes of program codes and providing more efficient information storage in the databases.

The efficiency of proposed concept was proven by the experiment, which showed that data storage in relational database using concept of MCIC is more efficient in contrast to usage of SCIC and HC. Using obtained results of measurements (see Tab.~\ref{tab-2}), dependencies between sizes of DBs and quantities of objects, which they contain (see Fig.~\ref{fig-1}) and dependencies between sizes of exported *.sql files of DBs and quantities of objects within DBs (see Fig.~\ref{fig-2}), were built. In addition, the method for calculation of approximate efficiency coefficient of data storage in relational database, using concept of MCIC, was proposed in Th.~\ref{MCIC_efficiency}.

However, despite all noted advantages of proposed extension of class notion, it requires further research, at least in the following directions:
\begin{itemize}
 \item comparison structure of MCICs with HCs hierarchies obtained using single and multiple inheritance;
 \item adoption of inheritance mechanisms for MCICs;
 \item building of inhomogeneous poly-hierarchies of MCICs;
 \item generalization of concept of MCICs to fuzzy case;
 \item adoption of inheritance mechanisms for fuzzy MCICs;
 \item building of inhomogeneous poly-hierarchies of fuzzy MCICs;
 \item studying of object-relation mapping of MCICs into:
 \begin{itemize}
  \item object-oriented databases;
  \item fuzzy object-oriented databases;
  \item graph databases;
 \end{itemize}
 \item adoption and usage of MCICs in object-oriented programming paradigm.
\end{itemize}

%
%

%
%

\end{document}